# PREDICTING THE PATH OF AN OPEN SYSTEM

S.Z.Stefanov*

NEK,National Dispatching, 8 Triaditsa Str.,

1040 Sofia, Bulgaria

The expected path of an open system, which is a big Poincaré system, has been found in this paper. This path has been obtained from the actual and from the expected droop of the open system. The actual droop has been reconstructed from the variations in the power and in the frequency of the open system. The expected droop has been found as a function of rotation from the expected potential energy of the open system under synchronization of that system. The expected path of an open system with resonances has been found by juxtaposing a molecule to the open system. The expected path of an open system with correlations has been found by juxtaposing a DNA-model to the open system. The expected path of an open system with balance has been found by juxtaposing radiation to the open system.

---

* E-mail: szstefanov@ndc.bg

## 1. INTRODUCTION

An open thermodynamic system exchanges energy or information with its environment. This system obtains energy or information from an external source which cannot be controlled or observed directly and without error (Mesarovic and Takahara,1975). The incoming information leads to a discrepancy between the system response and the system description. This discrepancy is observed (Mesarovic and Takahara,1975) as an unpredictable system response.

Let the open system be a big Poincaré system. Then, it is a non-integrable system with a continuous spectrum, according to the definition (Prigogine,1996) of a big Poincaré system. Under such consideration, the open system is characterized by continuous potential energy. Resonances and correlations are observed in this open system.

The resonances and correlations in an open system generate a transition to a new balance between the open system and the environment. The transition to a new balance is a process of establishing balance between the generation of power by the open system and the consumption of power by the environment under a new, lower frequency of the whole system. This process is characterized by the slope of the open-system static frequency characteristic, $k=\Delta P/\Delta f$. This characteristic corresponds to the droop of an electric power system (Ornov and Rabinovich,1988) and that is why we shall use the same term.

The expected transition to a new balance of powers sets a new potential energy of the open system and, therefore, determines the expected path of the open system. The aim of this paper is to pre-dict the path of an open system with resonances and correlations from the open-system droop.

## 2. SYNCHRONIZATION IN AN OPEN SYSTEM

Synchronization means establishing a sampling rate in a stationary system or stationarization of a system which has a sampling rate. The establishment of a sampling rate in a system makes sampling of the system possible.

The synchronization in an open system gives two open-system droops. Indeed, from the inverse proportionality of frequency and time, $f = 1/\tau$ , it follows that

$$k = \Delta P/\Delta f = -(\Delta P/\Delta \tau)\Delta \tau^2 \qquad (1)$$

That is why the synchronization, as stationarization with respect to power, $\Delta P/\Delta f = 1$ , gives

$$k^f = -\Delta \tau^2 \qquad (2)$$

Also, the synchronization, as sampling at $\Delta \tau = 1$ , gives

$$k^P = \Delta P/\Delta \tau \qquad (3)$$

The coefficients $k^f$ and $k^P$ are the two droops, set by the synchronization in an open system.

The optimal linear filter is (Sage and Melsa,1971) an algorithm of stochastic approximation. We shall accept that the optimal filter for finding an optimal linear estimate of the power P of an open system, under a given observation of the system's potential energy V, is a step from an algorithm of stochastic approximation

$$\Delta P = -[\ \Delta V\ ]_0 \qquad (4)$$

In (4), $[\Delta V]_0$ is norming of $\Delta V$ in the interval ( $\pi/2$, $\pi$ ] by means of dividing by $2^n$ for a suitably chosen integer n.

Indeed, (4) sets a step from an algorithm of stochastic approximation. This is so because, under open-system synchronization, $\Delta V/\Delta P = \Delta V$ is fulfilled.

The stochastic approximation, related to (4), involves the operation of norming $[\bullet]_0$ , and not the operation of sign($\bullet$). This is so because the norming $[\bullet]_0$ is a topological analogue of sign($\bullet$) in the zero vicinity of the function V(P) under a non-loaded electromechanical system with three generators, which system has a stable state of equilibrium. The non-loaded electromechanical system with three generators has a stable state of equilibrium (Vainman,1981), if one and only one arc of a circumference, out of the three arcs corresponding to the three generators, is bigger than $\pi/2$ . Therefore, the stochastic approximation under (4) has the norming $[\bullet]_0$ .

From (2), (3), (4), it follows that the droops $k^f$ and $k^p$ are obtained from the open-system potential energy thus

$$k^f = [\Delta\tau^2]_0, \ k^p = [\Delta V/\Delta\tau]_0 \tag{5}$$

From (5), it follows that the droops are given by the open-system rotation which is set by $[\bullet]_0$ .

## 3. EXPECTED PATH OF AN OPEN SYSTEM

Let the droop $k^f$ be a potential of a natural system

$$c'' = k^f \tag{6}$$

Here the natural system has one degree of freedom, it is set on a circumference, and it is assumed to have a potential with only non-degenerate singular points. Then, this natural system is of the kind as in (Selivanova,1998).

Since the open system is synchronized, then its sampling leads to fixing the length of the path of the natural system (1) in the following way

$$c'(0) = L_f \qquad (7)$$

Since the open system is synchronized, then its stationarization determines the following discrete dynamic system

$$P_{n+1} = P_n - u_n \qquad (8)$$

In (8), n is a natural number, and $u_n$ is control.

For this system (8), let the following problem be solved by the method of dynamic programming

$$\min_{u_0, u_1, \ldots, u_{N-1}} \quad \sum_{i=0}^{N-1} \hat{c}(P_i, u_i) + \delta_0(P_N) \qquad (9)$$

In (9), $\delta_0(P_N) = +\infty$ for $P_N \neq 0$ and $\delta_0(P_N) = 0$ for $P_N = 0$ . The function $\hat{c}$ from (9) is obtained from c through Fenchel's reverse transform.

From the asymptotic theorems of dynamic programming (Quadrat, 1990), applied to the problem under consideration, it follows that

$$c''(1) = \exp(2L_f)/c''(0) \qquad (10)$$

In (10), the splitting of time into future and past is taken into account when predicting the path of an open system. Due to this splitting of time in two, $c''(1)$ is present in (10).

From (10), the following estimate of the predicted path is obtained

$$L_f = 0.5(\ln(k^f_{t-1}) + \ln(k^f_t)) \tag{11}$$

In (11), $k^f_{t-1}$ is the actual droop for yesterday, and $k^f_t$ is the expected droop for today.

Analogously, the following estimate of the predicted path is obtained for $k^p$

$$L_p = 0.5(\ln(k^p_{t-1}) + \ln(k^p_t)) \tag{12}$$

The length of the total expected path of the open system is an Euclidean length or a ropelength (Baranska et al.,2005), (Mizel and Lidar,2004)

$$L = (L_f^2 + L_p^2)^{1/2} \quad \text{or} \quad (L_f + 4L_p)/5 \tag{13}$$

## 4. DETERMINATION OF THE ACTUAL DROOP

The failures found and estimated in the frequency for yesterday give the change in the frequency for yesterday, $\Delta f(t)$, $t=1,…,T$. This change sets a path x

$$X = k_0 t \sum_{i=1}^{T} \Delta f(i), \quad k_0 = \text{const} \tag{14}$$

Then the frequency spectrum f(x) is found as a solution of the following Fredholm integral equation of the first kind

$$\int_0^1 [1-x/t]_+ f(x)\,dt = f(t) \tag{15}$$

In (15), x is set by (14), and $[\bullet]_+ = (\bullet)$ if $(\bullet)\geq0$ and $[\bullet] = 0$ if $(\bullet)<0$.

From (15), it follows that the frequency spectrum is reconstructed from the behaviour of the frequency in time. This reconstruction has the meaning of reconstruction under nuclear spectro-scopy, according to the results for nuclear spectroscopy in (Vapnik, 1979).The reconstructed frequency spectrum will be denoted by $f^*(x)$.

The difference between the predicted and the actual power for yesterday sets the change in the power for yesterday. The power spectrum is reconstructed from the actual behaviour of the power in time under a given change in the power. This occurs in a manner analogous to that for the frequency. The reconstructed power spectrum will be denoted by $P^*(x)$.

The frequency spectrum and the power spectrum for x=T set $\Delta f_{t-1}$ and $\Delta P_{t-1}$

$$\Delta f_{t-1} = f^*(T) - 1, \ \Delta P_{t-1} = P^*(T) - 1 \tag{16}$$

Then, the actual droop is

$$k^f_{t-1} = k^P_{t-1} = \Delta P_{t-1}/\Delta f_{t-1} \tag{17}$$

## 5. SAMPLING UNDER RESONANCES

Juxtaposing an electric power system with impulse failure to a non-ergodic open system sets (Stefanov,2001) an self-simular behaviour of the open system. Under this juxtaposing, the non-integrability of the open system is neglected.

The oscillation coefficients of this electric power system are obtained from the invariants $g_2$ and $g_3$ of the Weierstrass function with frequencies $\omega_1$ and $\omega_2$, which corresponds to the open system.

Neglecting the non-integrability under this juxtaposing is due to neglecting the torsion in the open-system behaviour. This torsion is characterized by the Quillen metric from (Bost,1991). The torsion makes sampling of the open-system behaviour. That is why, the droop $k^f_{t,1}$ will be defined as follows

$$k^f_{t,1} = [ \; |\Delta|^{-1/12} \; ]_0, \; \Delta = g_2^3 - 27g_3^2 \tag{18}$$

The expected path $L_{f,1}$ for this droop and for $k^f_{t-1}$ from (17) is

$$L_{f,1} = 0.5(\ln(k^f_{t-1}) + \ln(k^f_{t,1})) \tag{19}$$

The internal potential energy of an open system, $V^{in}$ , will be set as follows

$$V^{in} = |\Delta|^{-1/12} \tag{20}$$

Then, $V^{in}$ can be considered as internal potential energy of a two-atomic molecule with inter-atomic distance $|\Delta|$. Indeed (Domenicano and Hargittai,1992), the internal potential energy of a two-atomic molecule with inter-atomic distance $|\Delta|$ can be $|\Delta|^{-1/12}$ .

### 6. STATIONARIZATION UNDER RESONANCES

Juxtaposing a thin wing to a non-integrable open system also sets (Stefanov, 2001) an self-simular behaviour of the open system. Then the open-system movement is characterized by a horizontal speed $u_i, i=1,…,T$, and a vertical speed $v_i, i=1,…,T$, which speeds are set by the horizontal and the vertical speed of streamlining a thin wing.

At every moment i, i=1,…T, this movement of the open system generates a travelling wave, modulated with respect to amplitude. The synchronization of the clocks of the open-system moving wing and of a stationary observer will be set as the following relation between the wing's own time $\Delta\tau$ and the time $\Delta t=1$ of the stationary observer

$$\Delta\tau^2 = \sum_{i=1}^{T} (1-(u_i/v_i)^2) \tag{21}$$

This synchronization is determined (Propoy,1998) as training under a non-local search for extremum in a wave medium. That is why, the synchronization of clocks sets a stationarization of the open system. Then, the droop $k^P_{t,1}$ will be defined in the following way

$$k^P_{t,1} = [ \ \Delta\tau^2 \ ]_0 \tag{22}$$

In (22), $\Delta\tau^2$ is set by (21).

The expected path $L_{p,1}$ for this droop and for $k^P_{t-1}$ from (17) is

$$L_{p,1} = 0.5(\ln(k^P_{t-1}) + \ln(k^P_{t,1})) \tag{23}$$

The external potential energy of an open system, $V^{out}$ , will be set as follows

$$V^{out} = \Delta\tau^2 \tag{24}$$

Then, $V^{out}$ can be considered as external potential energy of a two-atomic molecule with inter-atomic distance $\Delta\tau$ . Indeed (Domenicano and Hargittai,1992), the external potential energy of a two-atomic molecule with inter-atomic distance $\Delta\tau$ can be $\Delta\tau^2$ .

## 7. SAMPLING UNDER CORRELATIONS

Let the open system have four frequencies $\omega_i, i=1,\ldots,4$, which form two pairs of Poincaré resonances

$$n_1\omega_1 + m_1\omega_2 = 0 \tag{25}$$

$$n_2\omega_3 + m_2\omega_4 = 0$$

We shall assume that the frequencies $\omega_i, i=1,\ldots,4$, are generated by correlations in the open system, which correlations can be presented as correlation in a Heisenberg ferromagnetic system with four particles. For the spectral presentation of a four-particle

Green function of a Heisenberg ferromagnetic system with frequencies $\omega_i$, i=1,…,4, the following is fulfilled (Iziumov and Skriabin,1987)

$$i(\omega_1 + \omega_2) = -i(\omega_3 + \omega_4) \qquad (26)$$

$$i(\omega_1 + \omega_4) = -i(\omega_2 + \omega_3)$$

$$i(\omega_2 + \omega_4) = -i(\omega_1 + \omega_3)$$

In (26), i is the complex unity.

The correlation in this open system will be defined thus

$$\rho_{3,1} = (\omega_1 + \omega_3)/(\omega_2 + \omega_4) \qquad (27)$$

The correlation from (27) can be considered as a function of rotation of a natural system with two degrees of freedom, on a tor, in the absence of self-crossing of its geodesic. Such a consideration is possible due to the results of Selivanova (Selivanova,1998) about the function of rotation of a natural system.

From (25) and (27), it follows that $\rho_{3,1}$ is a median of $\rho_1 = \omega_1/\omega_2$ and $\rho_2 = \omega_3/\omega_4$, when $\rho_1$ and $\rho_2$ are positive fractions and $\rho_1<\rho_2$. Then, the following is fulfilled

$$\rho_1 < \rho_{3,1} < \rho_2 \qquad (28)$$

Since $\rho_{3,1}$ is a median, then the open system has sampling. We shall accept that the droop $k^f_{t,2}$ is

$$k^f_{t,2} = [\ \rho_{3,1}\ ]_0 \qquad (29)$$

The expected path $L_{f,2}$ for this droop and for $k^f_{t-1}$ from (17) is

$$L_{f,2} = 0.5(\ln(k^f_{t-1}) + \ln(k^f_{t,2})) \qquad (30)$$

The droop $k^f_{t,2}$ , under the above consideration of the correlation in an open system, has the meaning of twisting of a circular DNA from (Waterman,1989).

## 8. STATIONARIZATION UNDER CORRELATIONS

Let the open system with correlations be modelled as follows

$$V' = ( V^{out} - V^{in} )/2 \qquad (31)$$

This open-system model is a Mesarovic global model from (Egorov,1980).

The correlation in this open system is

$$\rho_{3,2} = 1/( V^{out} - V^{in} ) \qquad (32)$$

The correlation (32) can be considered as a function of rotation of a natural system with two degrees of freedom, on a tor, with self-crossing of its geodesic. Such a consideration follows from the results in (Selivanova,1998).

From (31) and (32), it follows that the following is fulfilled

$$V' = 1/(2\rho_{3,2}) \qquad (33)$$

The open system is stationarized under correlations, because (33) holds true. That is why we shall accept that the droop $k^P_{t,2}$ is

$$k^P_{t,2} = [ \ \rho_{3,2} \ ]_0 \qquad (34)$$

The expected path $L_{p,2}$ for this droop and for $k^P_{t-1}$ from (17) is

$$L_{p,2} = 0.5(\ln(k^P_{t-1}) + \ln(k^P_{t,2})) \qquad (35)$$

The droop $k^P_{t,2}$ , under the above consideration of the correlation in an open system, has the meaning of rising of a circular DNA from (Waterman,1989).

## 9. PRODUCTION OF REDUNDANCY BY AN OPEN SYSTEM

Let the open-system spectrum consist of five colours $c_i \in [0,1)$, $i=1,…,5$. Also, let the colours $c_i, i=1,…,4$, present almost completely the open-system energy. Let $c_E$ be the colour of the open-system energy. This colour $c_E$ corresponds to the number of paths of the open system, with length fixed by the system's potential energy. This number has a Poisson distribution, according to the enumeration asymptotic results from the combinatorics in (Bender,1974).

That is why the open-system redundancy at moment τ=0 is set by $c_E$ and it is

$$R(0) = c_E^*, \quad c_E^* = PREG(c_1,c_2,c_3,c_4) \tag{36}$$

In (36), PREG is a Poisson regression. Here R(0) has a Poisson distribution and R(0) is estimated as a linear function of the colours $c_i, i=1,…4$.

The production of redundancy by the open system is set by the colour $c_5$ and it is

$$\Delta R = 8\cos(\pi c_5) \tag{37}$$

In (37) it is assumed that the colour values are eight. Here ΔR is set through the wave length $\lambda_5 = \pi c_5$.

Let the open-system redundancy at moment τ=1 be

$$R(1) = R(0) + \Delta R \tag{38}$$

From (38), it follows that R(1) is set by the first two terms of the Taylor series expansion of R(1).

The redundancy at moment τ=1 is obtained from (36), (37), (38) and it is

$$R(1) = c_E^* + 8\cos(\pi c_5) \tag{39}$$

The open-system wave length $\lambda_5$ sets sampling of the open system. That is why the droop $k^f_{t,3}$ will be defined as follows

$$k^f_{t,3} = [\ R(1)\ ]_0 \tag{40}$$

The expected path $L_{f,3}$ for this droop and for $k^f_{t-1}$ from (17) is

$$L_{f,3} = 0.5(\ln(k^f_{t-1}) + \ln(k^f_{t,3})) \tag{41}$$

## 10. PRODUCTION OF ENTROPY BY AN OPEN SYSTEM

Let us consider the open system as a source of spherical waves. Let the spectrum of this source be observed in the distant zone. The receiver, thus situated, registers resonance in the open system. Resonance is viewed here as a signal with a narrow frequency band.

Let this receiver have two channels whose radius-vectors are

$$R_1 = \pi c_1/8, \ R_2 = \pi c_2, \ R_1 > R_2 \tag{42}$$

In (42), $c_1$ and $c_2$ are the two colours of the open system. They correspond to open-system waves of lengths $\lambda_1 = \pi c_1$ and $\lambda_2 = \pi c_2$.

Let this receiver scan the open-system spectrum at speed $v_0$. Then (Guliaev et al.,1986), it recognizes the two lines in the open-system spectrum, which lines are set by the two colours $c_1$ and $c_2$.

Let this receiver be an ideal receiver for the open system. Then it registers the following open-system entropy at moment $\tau=0$

$$S(0) = \pi R_1^2/4 \tag{43}$$

Indeed, if the receiver is an ideal one, then it fully absorbs the waves from the source and is a black hole for these waves. From the results of (Hawking and Penrose,1996) about the entropy of a black hole, (43) follows.

Let the open-system entropy at moment $\tau=1$ be

$$S(1) = S(0) + \Delta S \tag{44}$$

From (44), it follows that $S(1)$ is set by the first two terms of the Taylor series expansion of $S(1)$.

The production of entropy by the open system, $\Delta S$ , is given by the area of the receiver's truncated cone

$$\Delta S = \pi(R_1{}^2 - R_2{}^2)(1+(tg(v_0))^2)^{1/2}/16 \tag{45}$$

Then, the open-system entropy at moment $\tau = 1$ is

$$S(1) = \pi R_1{}^2/4 + \pi(R_1{}^2 - R_2{}^2)(1+(tg(v_0))^2)^{1/2}/16 \tag{46}$$

The radius-vector of the open-system waves is

$$R^* = S(1)^{1/2} \tag{47}$$

This radius-vector sets stationarization of the open system. That is why the droop $k^P{}_{t,3}$ will be defined as follows

$$k^P{}_{t,3} = [\ R^*\ ]_0 \tag{48}$$

The expected path $L_{p,3}$ for this droop and for $k^P{}_{t-1}$ from (17) is

$$L_{p,3} = 0.5(\ln(k^P{}_{t-1}) + \ln(k^P{}_{t,3})) \tag{49}$$

## 11. TOTAL PATH OF AN OPEN SYSTEM

The expected total path of an open system with resonances is set by (13), (19), (23) and it is

$$L_m = (L_{f,1}{}^2 + L_{p,1}{}^2)^{1/2} \tag{50}$$

This expected total path can be considered as inter-atomic distance in a two-atomic molecule with $V^{in}$ from (20) and $V^{out}$ from (24).

The expected total path of an open system with correlations is set by (13), (30), (35) and it is

$$L_d = (L_{f,2} + 4L_{p,2})/5 \tag{51}$$

This expected total path can be considered as the ropelength of a DNA-twisting and a DNA-rising, according to it.7 and it.8. Here the ropelength is set as in (Baranska et al.,2005).

The expected total path of an open system with balance is set by (13), (41), (49) and it is

$$L_b = (L_{f,3} + 4L_{p,3})/5 \qquad\qquad (52)$$

This expected total path can be considered as the length of a wave radiated by an open system with redundancy production $\Delta R$ from (37) and entropy production $\Delta S$ from (45). Here the length of a wave is set as in (Mizel and Lidar,2004).

## 12. CONCLUSION

The main result of this paper are the three obtained possible paths of an open system which is a big Poincaré system. These paths are obtained from the open-system droop. The actual droop is reconstructed from the variations in the power and in the frequency of the open system. The expected droop is defined as a function of rotation from the expected potential energy of the open system under synchronization of that system. The synchronization of the open system is introduced as sampling or as stationarization.

The stationarization under resonances is set by a global search, and the stationarization under correlations is set by a global model. The sampling under resonances is obtained from two waves, and the sampling under correlations is obtained from four particles. The production of entropy is defined as the registration of radiation by a black hole. The production of redundancy is defined as residual radiation.

The three expected paths of an open system are found by juxtaposing. The expected path of an open system with resonances is found by juxtaposing a molecule to the open system. The expected path of an open system with correlations is found by juxtaposing a DNA-model to the open system. The expected path of an open system with balance is found by juxtaposing radiation to the open system.